\documentclass{article}

\PassOptionsToPackage{numbers, compress}{natbib}
\usepackage[dblblindworkshop,final]{neurips_2026}
\workshoptitle{AI for Stochastic Dynamics}

\usepackage[utf8]{inputenc}
\usepackage[T1]{fontenc}
\usepackage[hidelinks]{hyperref}
\usepackage{url}
\usepackage{booktabs}
\usepackage{amsfonts}
\usepackage{amsmath}
\usepackage{amssymb}
\usepackage{nicefrac}
\usepackage{microtype}
\usepackage{xcolor}
\usepackage{enumitem}
\usepackage{graphicx}

\newcommand{\dof}{\mathrm{dof}}
\newcommand{\dofref}{\dof_{\mathrm{ref}}}
\newcommand{\dofsamp}{\dof_{\mathrm{samp}}}
\newcommand{\tr}{\mathrm{tr}}
\newcommand{\E}{\mathbb{E}}
\newcommand{\Esamp}{\E_{\mathrm{samp}}}
\newcommand{\rhores}{\rho_{\mathrm{res}}}

\title{Measuring Effective Data Resolution\\ in Guided Diffusion Posteriors}

\author{%
  Ridham Patel \\
  Indian Institute of Technology Gandhinagar \\
  Gandhinagar, India \\
  \texttt{ridham.patel@iitgn.ac.in}
  \And
  Defu Cao \\
  University of Southern California \\
  Los Angeles, USA\\
  \texttt{defucao@usc.edu}
  \And
  Jiacheng Pang \\
  University of Southern California \\
  Los Angeles, USA\\
  \texttt{pangj@usc.edu}
  \And
  Yan Liu \\
  University of Southern California \\
  Los Angeles, USA\\
  \texttt{yanliu.cs@usc.edu}
}

\begin{document}

\maketitle

\begin{abstract}
Guided diffusion samplers are increasingly used to reconstruct physical fields from sparse
observations, but standard diagnostics do not say how much of the reconstruction was
actually determined by the data. We introduce effective data resolution for black-box
generative posteriors: a comparison between the resolution warranted by the inverse
problem, $\dofref$, and the resolution realised by the sampler, $\dofsamp$. A perturbation
estimator measures $\dofsamp$ and the spatial map $R(x,x)$ from sampler queries alone. We
validate the estimator against exact references and use it to study guided diffusion. The
resulting measurements show that guidance weight can strongly alter apparent information
transfer, that mean, spread and resolution are not jointly corrected by one weight even
with an exact prior and score, and that resolution fidelity does not follow reliably from
the apparent principledness of a guidance rule.
\end{abstract}

\section{Introduction}

Reconstructing a physical field from a handful of sensors is a canonical ill-posed inverse
problem~\citep{stuart2010inverse,tarantola2005inverse}, and generative modelling has become
a standard tool for it. The recipe is to train
a diffusion prior over solution fields and then steer sampling toward the observations
through a likelihood term, yielding full-field reconstructions with uncertainty from sparse
data. DiffusionPDE~\citep{huang2024diffusionpde}, score-based inverse-problem
samplers~\citep{song2022scoreinverse,chung2023dps,kawar2022ddrm}, score-based data
assimilation~\citep{rozet2023sda} and function-space variants~\citep{shi2025fundiff} are
all instances of it.

Such a reconstruction is usually assessed by how close it lands to the truth and by whether
its uncertainty is calibrated~\citep{gneiting2007scoring,kuleshov2018calibrated}. These
metrics do not, by themselves, identify the thing a scientist needs to know
before drawing a conclusion from the reconstructed field: which parts of it the
sensors actually determined. A sampler can produce accurate-looking, well-calibrated
output while the structure a reader finds interesting is supplied almost entirely by the
prior, or, in the other direction, while it commits to detail far beyond what the noise
justifies~\citep{stuart2010inverse}. Conventional diagnostics score the reconstruction and
its marginals, not the dependence of the reconstruction on the data.

\paragraph{Two quantities, one comparison.}
The distinction we make precise is between the information the inverse problem
warrants and the information the sampler realises.

The first is a property of the problem. Given a prior, a forward operator, a noise level
and a sensor layout, there is a classical answer to how many directions of the unknown
field the data can constrain: the resolution, or degrees-of-freedom-for-signal, of the
inverse problem \citep{backus1968resolution,rodgers2000inverse,stuart2010inverse}. With $n$
point sensors it is at most $n$, and in an ill-posed problem substantially less, because
the forward operator destroys information before the sensors see it. It is computable in
closed form from a Gauss--Newton Hessian without running any sampler. We write it $\dofref$
and claim no novelty for it; we use it as a reference standard.

The second is a property of the algorithm. A guided sampler at a particular guidance weight
implements some map from observations to reconstruction, $y \mapsto \Esamp[u \mid y]$, and
how strongly that map responds to the data is a fact about the sampler, not about the
inverse problem. We call it the sampler's effective data resolution, $\dofsamp$. It
need not equal $\dofref$, it is not currently reported, and, as we show, it is largely set
by the guidance weight; Appendix~\ref{app:sweep} documents this for the DiffusionPDE
configuration we inspected.

What we propose measuring is their ratio, the resolution-fidelity ratio
\begin{equation}
\rhores \;=\; \dofsamp/\dofref ,
\label{eq:rho}
\end{equation}
read as follows: near $1$ the sampler attributes about as much resolution to the
observations as they warrant; below $1$ it attributes less, so the reconstruction leans more
on the prior than it appears to; above $1$ it attributes more than the noise justifies.

\paragraph{Why the sampler side is hard.}
For an estimator with an explicit Hessian, resolution is a matrix one differentiates. A
guided diffusion posterior has neither a Hessian nor a closed form, can only be queried by
running it, and is stochastic. The measurement therefore has to treat the sampler as a
black box. We adapt the perturbation estimator for generalized degrees of freedom
\citep{ye1998gdf,efron2004estimation}, which perturbs the data and measures the response of
the fit, combining it with stochastic trace estimation
\citep{hutchinson1990trace,ramani2008mcsure}. The resulting estimator uses two evaluations of the sampler's
reconstruction map per probe, a count that does not grow with the field dimension, and it
also yields the spatial map. The scalar says how much information the data appear to have
supplied; the map says where that apparent influence lands.

\paragraph{What the measurement reveals.}
We apply it to guided diffusion in two main settings and one appendix breadth check. In a
1D backward-heat problem with a
Gaussian prior the diffused score and the true posterior are both available in closed form,
so training error and prior mismatch are eliminated by construction and every deviation is
attributable to the guidance scheme. There, the guidance weight moves effective resolution
over $0.15$ to $1.27$ times the warranted value, and no single weight gets posterior mean,
spread and resolution right together. In a 2D nonlinear Darcy problem with a non-Gaussian
prior and a trained score, a comparable span reappears, but the advantage that principled
guidance held in the controlled setting does not carry over. A non-PDE compressed-sensing
check in Appendix~\ref{app:extra} is included only as a modest breadth check: it confirms
that the measurement is not tied to PDE forward operators, while leaving image restoration,
MRI, CT and other application domains to future work.

\paragraph{Contributions.}
\begin{enumerate}[leftmargin=1.3em,itemsep=1pt,topsep=2pt]
\item \textbf{A measurement of effective data resolution.} We define $\dofsamp$ for
  black-box generative posteriors, pair it with the classical reference $\dofref$, and
  estimate the fidelity ratio $\rhores$ and spatial map $R(x,x)$ from sampler queries
  (\S\ref{sec:resolution}, \S\ref{sec:estimator}).
\item \textbf{Validation and operating conditions.} We certify the estimator against exact
  references (bias $-0.009\pm0.028$, standard error $1/\sqrt{P}$), test it on a stochastic
  sampler, and identify the reproducibility, mixing and perturbation-scale conditions needed
  for trustworthy measurements (\S\ref{sec:validation}, \S\ref{sec:conditions}).
\item \textbf{Empirical analysis of guided diffusion.} In a controlled 1D problem, guidance
  weight moves $\rhores$ over $0.15$ to $1.27$ and no weight is jointly correct for mean,
  spread and resolution; in nonlinear 2D Darcy, a comparable range appears but the controlled
  method ordering does not persist; a non-PDE compressed-sensing check gives a modest
  breadth test of the same separation between resolution and conventional metrics
  (\S\ref{sec:1d}, \S\ref{sec:2d}; Appendix~\ref{app:extra}).
\end{enumerate}

\section{Related work}

\paragraph{Guided diffusion for PDE inverse problems.}
DiffusionPDE~\citep{huang2024diffusionpde} learns a joint distribution over coefficient and
solution fields and guides sampling with sparse observations together with a PDE residual.
Score-based methods solve inverse problems by steering a learned prior at test time
\citep{song2022scoreinverse,chung2023dps,kawar2022ddrm}; score-based data
assimilation~\citep{rozet2023sda} decouples the observation model from an
unconditional trajectory prior, and function-space
formulations~\citep{shi2025fundiff} target discretisation invariance. All of them steer an
unconditional prior with an observation term, and all inherit an approximation in doing so,
because the exact guidance term $\nabla_{x_t}\log p(y \mid x_t)$ is intractable in general.
Diffusion
posterior sampling (DPS)~\citep{chung2023dps} replaces it with
$\nabla_{x_t}\log p(y \mid \hat{x}_0(x_t))$, evaluating the likelihood at the denoised
estimate. This substitutes a surrogate likelihood gradient for the intermediate-time
likelihood gradient; its effect on how much information the sampler transfers has not, to
our knowledge, been quantified.
$\Pi$GDM~\citep{song2023pigdm} models the $x_0 \mid x_t$ covariance isotropically, and
DDNM~\citep{wang2023ddnm} and DDRM~\citep{kawar2022ddrm} motivate data-consistency
projection schemes. These form the approximation ladder that we use in the experiments.

\paragraph{Uncertainty quantification for neural PDE surrogates.}
Linearised neural operators yield function-valued Gaussian
processes~\citep{magnani2024luno}, and predictive calibration methods address reliability of
reported intervals~\citep{kuleshov2018calibrated};
large pretrained PDE models~\citep{herde2024poseidon} and neural
operators~\citep{li2021fno} raise the same questions at scale. These quantify
surrogate error: how wrong the learned operator is. Our question is different and
complementary, concerning inverse-problem resolution, namely how much the sensors
determined, which is a property of the data and the sampler rather than of the operator's
accuracy.

\paragraph{Resolution analysis and perturbation-based degrees of freedom.}
The estimand is classical. Resolution operators and averaging kernels go back to
\citet{backus1968resolution}, degrees of freedom for signal are standard in atmospheric
retrieval \citep{rodgers2000inverse}, and low-rank posterior structure is well studied in
large-scale Bayesian inverse problems
\citep{tarantola2005inverse,stuart2010inverse,flath2011lowrank,spantini2015optimal}. So is the estimator:
\citet{ye1998gdf} and \citet{efron2004estimation} define generalized degrees of freedom as
$\sum_i \partial \hat y_i / \partial y_i$ and estimate it by perturbing the data and
measuring the response of the fit. Monte-Carlo SURE uses the same black-box perturbation
idea for denoisers~\citep{ramani2008mcsure}, and Hutchinson probing
\citep{hutchinson1990trace} is the standard device for the trace.

Our contribution is not a new trace estimator. It is the transplant of this machinery to
posteriors that can only be sampled, which requires common random numbers, a
noise-scaled perturbation inside a window we characterise, and a smooth coupling condition;
the certification of the resulting measurement against exact references; and the pairing of
the measured quantity with a computable problem-side reference so that a fidelity ratio can
be formed. To our knowledge no formulation of resolution analysis has been applied to a
generative posterior, or used as a standard diagnostic for guided diffusion inverse
problems.

\section{Effective data resolution}
\label{sec:resolution}

\paragraph{Setting.}
Let $u \in \mathbb{R}^N$ be the unknown field on a discretised domain and let
\begin{equation}
y = Au + \eta \in \mathbb{R}^n, \qquad \eta \sim \mathcal{N}(0,\sigma_y^2 I), \qquad n \ll N,
\end{equation}
be the observations, where $A$ is the (linearised) forward operator mapping the field to
sensor readings and $\sigma_y$ is the noise level. An estimator is any map
$m : \mathbb{R}^n \to \mathbb{R}^N$ from data to reconstruction; for us $m$ will be the
posterior mean of a sampler~\citep{stuart2010inverse,tarantola2005inverse}.

\paragraph{The resolution operator, informally.}
Imagine changing the true field slightly at location $x'$ and asking how the reconstruction
at location $x$ responds. Collecting these sensitivities gives the resolution
operator
\begin{equation}
R \;=\; \frac{\partial m}{\partial u} \;=\; \frac{\partial m}{\partial y}\,A ,
\label{eq:R}
\end{equation}
an $N \times N$ object whose $(x,x')$ entry says how much a change in the truth at $x'$
propagates into the estimate at $x$. A perfect estimator that recovered the field exactly
would have $R = I$. In an ill-posed problem it cannot: the data do not determine the field,
so $R$ is a smoothing operator~\citep{backus1968resolution,rodgers2000inverse}.

Its diagonal $R(x,x)$ is the interpretable object. A value near $1$ means the estimate at
$x$ tracks the truth at $x$, so that location is determined by the data. A value near $0$
means the estimate is weakly determined by the data there, so prior information contributes
more strongly to what appears at that location. The diagonal is thus a map of where data
influence lands.

\paragraph{The scalar: effective data resolution.}
Summing the diagonal counts the number of independent directions the estimator lets the
data determine,
\begin{equation}
\dofsamp \;=\; \tr(R),
\label{eq:dofsamp}
\end{equation}
which we call the sampler's effective data resolution. It is a property of that
algorithm at that guidance weight, not of the inverse problem: two samplers targeting the
same posterior can, and do, return different values.

\paragraph{The reference: what the problem warrants.}
For the exact Bayesian posterior of a linear--Gaussian problem with prior covariance $C$,
the corresponding quantity is classical and requires no sampling. The prior-preconditioned
Gauss--Newton Hessian
\begin{equation}
H \;=\; C^{1/2}A^\top A\,C^{1/2}/\sigma_y^2
\end{equation}
has eigenvalues $\lambda_i$, each measuring how strongly the data constrain one
prior-whitened direction, and
\begin{equation}
\dofref \;=\; \sum_i \frac{\lambda_i}{1+\lambda_i}.
\label{eq:dofref}
\end{equation}
Each term lies in $[0,1)$ and is near $1$ for a well-constrained direction and near $0$ for
one the data do not see, so $\dofref$ is the information budget the observations carry.
Since $\mathrm{rank}(H) \le n$, it is cheap even at large $N$. For a nonlinear forward map
we linearise at the posterior mean, which is standard practice in retrieval
\citep{rodgers2000inverse,flath2011lowrank,spantini2015optimal}.

\paragraph{Resolution fidelity.}
The comparison we propose is the resolution-fidelity ratio $\rhores = \dofsamp/\dofref$ of
\eqref{eq:rho}. As a definition it says only this: $\rhores \approx 1$ means the sampler
attributes about as much resolution to the observations as the inverse problem warrants,
$\rhores < 1$ means it attributes less, and $\rhores > 1$ means it attributes more. Both
departures occur in our experiments.

The interpretation is a separate step, and in our settings it is borne out by the other
columns. A sampler with $\rhores < 1$ responds weakly to the data, so structure in its
reconstruction is less data-driven than it appears and is correspondingly more attributable
to the prior. A sampler with $\rhores > 1$ responds to the observations more strongly than
the noise justifies, which in \S\ref{sec:1d} coincides with variance ratios far above one
and coverage at $100\%$: detail that the measurements do not support.

\paragraph{Relation to existing diagnostics.}
Coverage, CRPS \citep{gneiting2007scoring} and posterior variance ratios score the
distribution a sampler reports: whether its intervals contain the truth at the advertised
rate, and whether its spread is right. Resolution is complementary rather than a
replacement: it differentiates the reconstruction map with respect to $y$, and therefore
measures data dependence that fixed-observation diagnostics do not directly access.

\section{Black-box estimation}
\label{sec:estimator}

Computing $R$ from \eqref{eq:R} directly is an $N \times N$ object and out of reach. The
reduction that makes the problem tractable is cyclicity of the trace,
\begin{equation}
\tr(R) \;=\; \tr\!\big(A\,\partial m/\partial y\big),
\label{eq:cyclic}
\end{equation}
where $A\,\partial m/\partial y$ is only $n \times n$: the estimand lives in observation
space, whose dimension is the sensor count, not in field space. We never form it. A
Rademacher probe $z \in \{\pm1\}^n$ gives the estimator
\begin{equation}
\tr(R) \;\approx\; \E_z\!\left[z^\top A\,\frac{m(y+\epsilon z)-m(y-\epsilon z)}{2\epsilon}\right],
\label{eq:hutch}
\end{equation}
averaged over $P$ independent probes~\citep{hutchinson1990trace,ramani2008mcsure}.
Throughout, $n$ denotes the sensor count and $P$ the probe count; they are independent
quantities.

\paragraph{Cost.}
Each probe requires two evaluations of the reconstruction map $m$, and the sampler is used
only as a black box. It is worth being precise about what that costs, because $m(y)$ is
itself the posterior mean of a stochastic sampler and must be approximated by an ensemble of
$E$ trajectories, so the total is $2PE$ trajectories. What does not grow with the field
dimension $N$ is the number of estimator evaluations, $2P$: the trace estimator
introduces no dimensional dependence of its own, and whatever cost $N$ imposes enters
through the price of a single sampler trajectory, exactly as it would if one were merely
drawing from the posterior.

\paragraph{The spatial map.}
Replacing the random probes by the $n$ coordinate directions of $y$ gives the columns of
$\partial m/\partial y$ exactly, and hence the diagonal $R(x,x)$, at $2n$ evaluations and
with no probe variance. This is affordable precisely in the sparse-sensor regime of
interest, and it is what turns the scalar into the map of \S\ref{sec:resolution}.

\paragraph{Common random numbers.}
The samplers we audit are stochastic, so $m(y)$ is estimated from a finite ensemble. The
two evaluations in \eqref{eq:hutch} must be run under common random numbers, that is, with
the same seed, so that the ensemble noise cancels rather than being divided by $2\epsilon$.
For a diffusion sampler this cancellation is exact rather than approximate: the noise
draws depend only on the seed and never on $y$, so for a fixed seed $m(\cdot)$ is a
deterministic function of the data~\citep{glasserman2004monte}.

\paragraph{Coupling requirement.}
The estimator therefore requires more than reproducibility at a fixed input: the two
perturbed runs must remain smoothly coupled as $y$ changes. Diffusion and flow samplers
satisfy this. Metropolis-corrected samplers such as pCN~\citep{cotter2013mcmc} can be
replayed from a seed, but their accept/reject discontinuities can split the two perturbed
paths, so chain noise swamps the $O(\epsilon)$ signal. We return to this in
\S\ref{sec:conditions}.

\paragraph{Perturbation scale.}
The step $\epsilon$ must be scaled to the observation noise $\sigma_y$ rather than set in
absolute units, and it must lie inside a window. Its upper edge is ordinary
finite-difference bias. It also has a lower edge for the residual-normalised DPS
implementation we study, where the normalisation makes the reconstruction map near-singular
at small scales; schemes without that normalisation show no such edge. We characterise the
window in \S\ref{sec:conditions} and Appendix~\ref{app:eps}, and report $\epsilon$ with
every measurement.

\section{Validation}
\label{sec:validation}

Before using the measurement to draw conclusions about guided diffusion, we establish that
the measurement itself is trustworthy. We do so on a backward-heat problem with a
$128$-dimensional field and $12$ sensors in a half-domain layout, where the posterior is
Gaussian in closed form and $\dofref = 5.971$ exactly.

The estimator passes three checks (Table~\ref{tab:cert}, Appendix~\ref{app:cert}). First, it
is unbiased against the exact reference: at $P=128$ probes the bias is
$-0.009\pm0.028$. Second, its standard error follows the ordinary Monte-Carlo rate
$1/\sqrt{P}$; increasing $P$ from $8$ to $128$ lowers the per-run standard deviation
$0.897\to0.179$, a factor $5.0$ against the $4.0$ predicted. Third, it works on a stochastic
but reproducible sampler: preconditioned unadjusted Langevin gives $6.33\pm0.59$ against the
exact $5.97$.

This last check also fixes the main prerequisite. Resolution can only be measured from a
sampler whose posterior mean is itself trustworthy. Without low-rank posterior
preconditioning~\citep{flath2011lowrank,spantini2015optimal}, the unadjusted Langevin
sampler~\citep{durmus2019ula} had
effective sample size about $40$ out of $150$k steps and under-dispersed by $2$ to
$8\times$; after preconditioning it reached ESS $13{,}753$ and matched the closed-form
posterior to $0.3\%$ in mean and $0.6\%$ in standard deviation. The perturbation scale is a
separate condition: reference samplers with affine-in-$y$ means are flat in $\epsilon$, but
guided samplers are not. A higher-replicate stress test in Appendix~\ref{app:extra}
confirms that the exact-score control remains stable while high-weight DPS becomes
realisation-sensitive at reduced probe budgets. We return to these conditions in
\S\ref{sec:conditions}.

\section{Guidance controls apparent information content}
\label{sec:1d}

\paragraph{Design, and why it isolates guidance.}
We take a 1D backward-heat problem: a $64$-dimensional field observed through $8$ point
sensors after diffusive smoothing, with $\dofref = 7.85$. The prior is Gaussian, so the
diffused score and the Tweedie denoiser~\citep{efron2011tweedie} are available in closed form
and the true posterior is known exactly. There is therefore zero training error and zero
prior mismatch, and any departure from the reference posterior is attributable to the
guidance scheme alone. This is what a released checkpoint cannot offer, since there guidance
error, score error and prior mismatch are confounded. We compare exact guidance as a control
arm against
$\Pi$GDM~\citep{song2023pigdm}, a data-consistency projection in the style of
DDNM~\citep{wang2023ddnm}, and DPS~\citep{chung2023dps} at several guidance weights $\zeta$.
All resolution estimates use $\epsilon = \sigma_y$.

\begin{table}[t]
\centering
\small
\caption{Guidance weight trades posterior accuracy, spread, and effective resolution in
the controlled 1D backward-heat problem. ``mean err'' and ``std err'' are relative
$L^2$ errors in the posterior mean and pointwise standard deviation; ``var ratio'' is
relative to the closed-form posterior, CRPS is the continuous ranked probability score,
and ``cov@90\%'' is empirical coverage of the nominal 90\% interval. The fidelity ratio
$\rhores=\dofsamp/\dofref$ is measured at $\epsilon=\sigma_y$ over $4$ ensemble
realisations; $\pm$ denotes the across-realisation standard error.}
\label{tab:1d}
\begin{tabular}{lrrrrrr}
\toprule
guidance & mean err & std err & var ratio & CRPS & cov@90\% & $\rhores$ \\
\midrule
exact score (control)      & 0.040 & 0.046 & 0.97 & 0.309 & 0.812 & \textbf{0.985 {\scriptsize$\pm$.002}} \\
$\Pi$GDM                   & 0.041 & 0.047 & 0.96 & 0.307 & 0.812 & 1.022 {\scriptsize$\pm$.002} \\
projection                 & 0.051 & 0.049 & 0.94 & 0.308 & 0.812 & 1.019 {\scriptsize$\pm$.000} \\
DPS $\zeta=0.3$            & 0.155 & 0.521 & 2.22 & 0.314 & 0.938 & 0.796 {\scriptsize$\pm$.038} \\
DPS $\zeta=1$              & 0.074 & 2.712 & 13.8 & 0.356 & \textbf{1.000} & 1.219 {\scriptsize$\pm$.046} \\
DPS $\zeta=3$              & 0.350 & 9.453 & 97.7 & 0.528 & \textbf{1.000} & 1.053 {\scriptsize$\pm$.129} \\
DPS, literal $1/\sigma_y^2$ & \multicolumn{6}{c}{diverged ($\max|x| \sim 10^{173}$)} \\
\bottomrule
\end{tabular}
\end{table}

\begin{figure}[t]
\centering
\includegraphics[width=\linewidth]{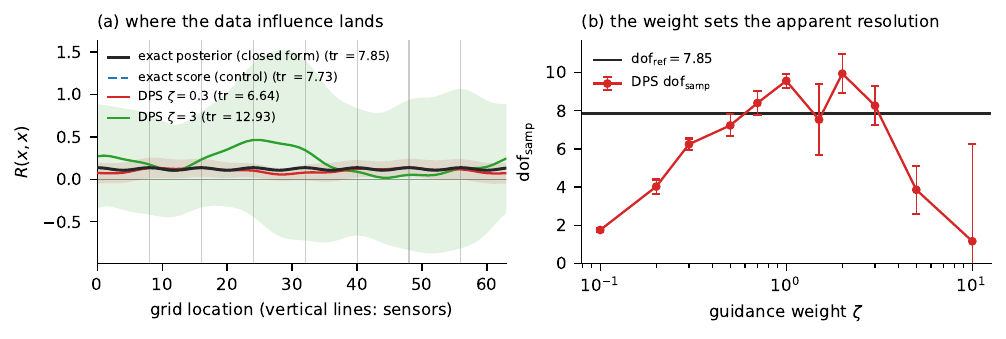}
\caption{Effective data resolution in the controlled 1D setting, at $\epsilon = \sigma_y$.
\textbf{(a)} The spatial map $R(x,x)$, averaged over $16$ ensemble realisations with
$\pm 1$ sd bands, showing where apparent data influence lands. The exact posterior
spreads it almost uniformly ($0.107$ to $0.138$) and the exact-score control recovers that
to $0.002$ with zero spread across realisations, which validates the apparatus. DPS at
$\zeta=0.3$ under-resolves roughly uniformly (averaged trace $6.64$ against $7.85$), while
DPS at $\zeta=3$ concentrates influence in a band near $x=24$ and starves the rest of the
field ($0.019$ to $0.464$); the latter's map is not even stable across realisations, its
per-realisation trace ranging from $-24.0$ to $68.2$. \textbf{(b)} The scalar $\dofsamp$
against guidance weight ($4$ realisations, error bars are across-realisation standard
errors), with $\dofref$ as the horizontal line. The curve rises, crosses the reference, and
falls back, so $\dofsamp = \dofref$ has more than one root and matching resolution does not
identify a guidance weight.}
\label{fig:localization}
\end{figure}

\paragraph{The control arm validates the apparatus.}
Exact guidance returns $0.985\,\dofref$, and $\Pi$GDM and projection land within $2\%$
(Table~\ref{tab:1d}). Since these schemes closely match the reference posterior across the reported diagnostics, a departure elsewhere in the table is a property of the guidance scheme rather than
of the measurement.

\paragraph{Guidance weight sets the apparent information content.}
Sweeping $\zeta \in [0.1,10]$ (Figure~\ref{fig:localization}b, Appendix~\ref{app:sweep})
moves $\dofsamp$ from $0.15$ to $1.27$ times $\dofref$: at $\zeta = 0.1$ the sampler behaves
as though the sensors had determined about a fifth of what they carry, near $\zeta = 2$ as
though they had determined more than the noise justifies, and by $\zeta = 10$ it falls back
to $0.15$. The same observations, the same prior and the same score thus produce an
eightfold difference in apparent information content, depending on a guidance weight
rather than on the inverse problem alone.
Thus, in this controlled setting, changing $\zeta$ substantially changes how much data dependence the sampler realises.

\paragraph{No single weight is jointly correct.}
Reading down the DPS rows of Table~\ref{tab:1d}: the weight with the best posterior mean
($\zeta = 1$, mean error $0.074$) carries $13.8\times$ the correct variance and reports
$100\%$ coverage at the nominal $90\%$ level; the weight with the best distributional match
($\zeta = 0.3$) resolves only $0.80\times$ the warranted degrees of freedom; and the weight
whose resolution is closest to correct ($\zeta = 3$, at $\rhores = 1.053$) has $97.7\times$
the correct variance and the worst CRPS in the table. No setting gets mean, spread and
resolution right together. This is not a failure of tuning that a finer sweep would repair,
because the sweep is what exhibits it, and it holds under the most favourable conditions
available.

Mechanistically, the DPS rows differ from the exact and $\Pi$GDM controls because the data
term is evaluated at the denoised estimate and then scaled by the residual norm. Changing
$\zeta$ therefore changes both the strength of the data pull and the region of the
trajectory where residual-normalised corrections dominate the unconditional diffusion
dynamics. Appendix~\ref{app:eps} isolates the numerical consequence: for this DPS
implementation the local derivative is singular as $\epsilon\to0$, so the reported
resolution is a finite-scale sensitivity, not an infinitesimal Jacobian.

\paragraph{Matching resolution is therefore not a tuning rule.}
Some $\zeta$ does hit $\dofref$, but it is not the one with the best mean or spread, and the
sweep is non-monotone, so $\dofsamp = \dofref$ has at least two roots, near
$\zeta \approx 0.6$ and $\zeta \approx 4$, belonging to very different samplers. The
measurement reports a discrepancy; it does not select a setting.

\paragraph{The scalar hides two spatial failures.}
Figure~\ref{fig:localization}(a) shows what the trace alone cannot. DPS at $\zeta = 3$
mis-localises data influence: the exact posterior is nearly uniform over $[0.107,0.138]$,
whereas DPS spans $[0.019,0.464]$ and concentrates influence near $x = 24$.
The map is also unstable: across $16$ ensemble realisations, $\tr(R)$ ranges from
$-24.0$ to $68.2$, whereas the exact-score control reproduces the closed-form map to
$0.002$ with zero spread. A sampler whose resolution map changes sign between runs is not one
whose reported uncertainty can be taken at face value. Coverage and CRPS in
Table~\ref{tab:1d} respond to the over-dispersion at $\zeta = 3$, but, being computed at
fixed $y$, cannot reveal where the influence lands or whether it moves between runs.

\paragraph{What this experiment establishes, and what it leaves open.}
Because the score and the prior are exact here, this experiment establishes that effective
resolution can change purely as a consequence of guidance, and within this controlled setting, distinguishes the schemes: exact
guidance, $\Pi$GDM and projection all recover $\dofref$ to within $2\%$, while DPS does not
produce a jointly correct posterior mean, spread and resolution at the weights studied. It
cannot tell us whether either conclusion survives once the idealisations are removed. That
is the next experiment.

\section{Does resolution behaviour transfer? A nonlinear test}
\label{sec:2d}

We therefore ask whether the behaviour of \S\ref{sec:1d}, and the relative standing of the
guidance schemes within it, persist once score error, prior non-Gaussianity and
forward-model nonlinearity are introduced together.

\paragraph{What becomes more realistic.}
We use a 2D Darcy flow problem: $24 \times 24 = 576$ field unknowns observed at $64$
sensors, a nonlinear forward map $-\nabla\!\cdot(e^m \nabla u) = f$, a non-Gaussian prior
$m = A\tanh(g/s)$ built over a latent Gaussian $g$ (field kurtosis $1.88$ against $3.0$ for
a Gaussian), and a trained $100$k-parameter score network with EDM
preconditioning~\citep{karras2022edm}. 
Relative to \S\ref{sec:1d} this adds nonlinearity, prior non-Gaussianity and score approximation simultaneously.

\paragraph{An exact reference survives.}
The latent construction is what keeps the comparison meaningful. Although the field prior is
non-Gaussian, it is a deterministic transform of a Gaussian latent, and preconditioned
Crank--Nicolson~\citep{cotter2013mcmc} in that latent space is Metropolis-corrected and
therefore exact for the true posterior. This gives a trustworthy reference ensemble;
$\dofref = 43.15$, i.e.\ $7.5\%$ of the $576$ field unknowns. Resolution estimates use
$P=24$ Rademacher probes and $\epsilon = 0.5\,\sigma_y$.

\begin{table}[t]
\centering
\small
\caption{Resolution fidelity under a nonlinear map with a trained score, against an exact
pCN reference. ``rel-$L^2$'' is the relative $L^2$ error of the posterior mean,
``std/ref'' is ensemble spread relative to the pCN reference, ``SW$\to$ref'' is the
sliced-Wasserstein distance to that reference, CRPS is the continuous ranked probability
score, and ``cov@90\%'' is empirical coverage of the nominal 90\% posterior interval.
The span of $\rhores$ is comparable to the controlled setting, but projection over-resolves
by $50\%$ while DPS at $\zeta=1$ is the closest listed method. $\Pi$GDM diverged in our
implementation, so we exclude it rather than claim a result about the method.}
\label{tab:2d}
\begin{tabular}{lrrrrrr}
\toprule
guidance & rel-$L^2$ & std/ref & SW$\to$ref & CRPS & cov@90\% & $\rhores$ \\
\midrule
pCN (reference)   & 0.689 & 1.000 & 0.000 & 0.211 & 0.774 & 1.000 \\
$\Pi$GDM          & \multicolumn{6}{c}{diverged (implementation; see text)} \\
projection        & 0.808 & 1.219 & 0.255 & 0.249 & 0.785 & \textbf{1.501} \\
DPS $\zeta=0.3$   & 0.911 & 1.644 & 0.341 & 0.283 & 0.851 & 0.152 \\
DPS $\zeta=1$     & 0.796 & 1.373 & 0.270 & 0.244 & 0.845 & \textbf{1.026} \\
DPS $\zeta=3$     & 0.799 & 1.290 & 0.252 & 0.245 & 0.812 & 1.372 \\
\bottomrule
\end{tabular}
\end{table}

The central phenomenon replicates: across DPS weights, $\rhores$ ranges from $0.15$ to
$1.37$, comparable to the $0.15$ to $1.27$ of \S\ref{sec:1d}. What does not transfer is the
relative standing of the guidance schemes. Projection was within $2\%$ of the warranted
resolution in the controlled setting, but here over-resolves by $50\%$; DPS at $\zeta=1$ is
the closest listed method at $\rhores=1.026$. Nor does over-dispersion transfer in the same
form:
coverage stays between $0.785$ and $0.851$ against the reference's $0.774$, and spread ratios
are $1.2$--$1.6$ rather than $13.8$--$97.7$.

The scope is deliberately narrow. $\Pi$GDM diverged in our nonlinear implementation, so
projection is the only principled scheme represented here; we therefore claim only that
projection does not retain its 1D advantage, and that a scheme's approximation structure did
not predict $\rhores$ across these settings. The projection result is not caused by the
stale-Jacobian approximation: refreshing $JJ^\top$ every $25,5,1$ steps gives
$\rhores=1.501,1.482,1.479$.

\section{Use and scope}
\label{sec:conditions}

The measurement is useful only when its operating conditions are made explicit. The
perturbation scale must be tied to the observation noise. In 2D Darcy, using the 1D
absolute value $\epsilon=10^{-2}$ perturbs the data by $30\times$ the noise and returns
$\dofsamp=10.0$ against a true $43.1$. Nor can $\epsilon$ always go to zero: for the
residual-normalised DPS implementation studied here, $\|y-A\hat x_0\|^{-2}$ makes the
small-scale derivative singular (Appendix~\ref{app:eps}). We therefore report $\epsilon$,
use $\epsilon\in[0.1,1]\sigma_y$, and make no claim about DPS formulations with different
scalings.

The estimator also requires a reproducible, smoothly coupled sampler under common random
numbers. Diffusion samplers satisfy this; Metropolis-corrected samplers such as
pCN~\citep{cotter2013mcmc} can be replayed from a seed, but accept/reject discontinuities
can split the two perturbed paths. For pCN we use the Gauss--Newton reference
\eqref{eq:dofref}, not black-box $\dofsamp$. Randomised traces are useful mainly when
observation dimension is large: the coordinate-wise map costs $2n$ evaluations and has no
probe variance, while our calibration suggests $P\approx12$ probes for $10\%$ accuracy and
$P\approx289$ for $2\%$.


Our scale is deliberate: the comparison requires a trustworthy $\dofref$, available
analytically in 1D and through latent pCN in 2D. Released checkpoints, image restoration,
MRI, CT, robotics, consistency models, flow matching, rectified flows and newer inverse
solvers are natural targets, but each requires a defensible reference resolution. Our
claims are therefore limited to the guidance schemes implemented here, with nonlinear
resolution interpreted as a posterior-mean linearisation and severe over-dispersion treated
as a practical measurability limit. Appendix~\ref{app:extra} reports a non-PDE
compressed-sensing check (Table~\ref{tab:compressed}) and a higher-replicate stability
stress test (Table~\ref{tab:stability}).

\section{Conclusion}

A generative posterior can produce accurate reconstructions and plausible uncertainty
statistics while transferring the wrong amount of information from the data. We made that
quantity measurable by comparing $\dofref$, the problem-side resolution budget, with
$\dofsamp$, the sampler-realised resolution, and by estimating both $\rhores$ and the map
$R(x,x)$ from black-box sampler queries. Applied to guided diffusion, the measurement shows
that guidance weight can move apparent information content by nearly an order of magnitude,
that no single weight fixes mean, spread and resolution together even with an exact score
and prior, and that the controlled method ordering need not persist under nonlinear learned
settings. Information fidelity is therefore something to measure, not something to infer
from the form of a guidance derivation.

\bibliographystyle{plainnat}
\bibliography{references}

\appendix

\section{The perturbation scale: window and singular behaviour}
\label{app:eps}

\paragraph{Why the certification of \S\ref{sec:validation} does not generalise.}
The estimate is flat across $\epsilon \in [10^{-5},10^{-1}]$ to twelve significant figures on
both reference samplers. Both have a posterior mean that is affine in $y$ at a fixed noise
realisation: the exact Gaussian posterior by construction, and preconditioned unadjusted
Langevin because its update is affine in $y$ at fixed noise. For an affine map the central
difference is exact at every $\epsilon$, so flatness there is guaranteed and carries no
information about guided samplers.

\paragraph{Common random numbers are exact, so this is bias and not noise.}
The sampler draws its noise from a generator seeded independently of $y$, so for a fixed
seed $\hat m(y)$ is a deterministic, smooth function of $y$ and the two evaluations in
\eqref{eq:hutch} share their randomness exactly. There is no Monte-Carlo residual being
amplified by $1/2\epsilon$. What varies between runs is which realisation $\hat m$ one
obtains, which is why we report across-realisation error bars throughout.

\paragraph{Residual-normalised DPS admits no $\epsilon \to 0$ limit.}
DPS Algorithm 1 divides its correction by $\|y - A\hat x_0\|^{2}$, giving $\hat m$
near-singular structure at small scales. The statements below concern that implementation;
other scalings of the guidance step may behave differently and we did not test them.
Measuring $\zeta = 1$ with ensembles of $512$:

\begin{table}[ht]
\centering\small
\begin{tabular}{lrrrrrr}
\toprule
$\epsilon/\sigma_y$ & $10^{-6}$ & $10^{-4}$ & $10^{-2}$ & $0.1$ & $1$ & $10$ \\
\midrule
$\dofsamp$ & $-5.25\times10^{6}$ & $4.27\times10^{4}$ & $462.2$ & $10.94$ & $10.91$ & $8.35$ \\
\bottomrule
\end{tabular}
\end{table}

Differentiating the sampler confirms it. The map is differentiable in $y$ by automatic
differentiation, so $\tr(R) = \tr(A\,\partial \hat m/\partial y)$ follows from $n$
reverse-mode passes with cotangent $A^\top e_k$, at no $\epsilon$ at all. On schemes without
residual normalisation this reproduces the finite difference exactly and is independent of
the realisation: exact score $7.733$, $\Pi$GDM $8.016$, projection $8.000$, each with zero
spread across seeds and across ensemble sizes from $128$ to $2048$. On every DPS setting it
returns values of order $10^{47}$, which is the correct answer to a question that should not
be asked of that sampler.

\paragraph{Choosing inside the window.}
The upper edge is ordinary secant bias and the lower edge is the singularity above. Across
four realisations in 1D, the bias--variance tradeoff places the operating point at
$\epsilon = \sigma_y$: at $0.1\,\sigma_y$ the estimate is essentially unbiased but the
across-realisation standard deviation is $8$ to $100$, while at $10\,\sigma_y$ the spread is
small but the value is biased low ($5.54$, $7.66$, $7.88$ for $\zeta = 0.3,1,3$ against
$6.25$, $9.57$, $8.27$ at $\sigma_y$). All 1D results use $\epsilon = \sigma_y$. All 2D
results use $\epsilon = 0.5\,\sigma_y$, where the low-$\zeta$ estimate is stable ($6.650$ at
$0.1\,\sigma_y$ against $6.578$ at $0.5\,\sigma_y$ for $\zeta = 0.3$); at larger $\zeta$ the
2D estimates retain some probe-scale sensitivity, so the 2D values in Table~\ref{tab:2d}
should be read as accurate to their leading digit rather than to three.

\section{Three formulations that do not work}
\label{app:dead}

Three natural-looking alternatives to $\tr(R)$ are degenerate, one of them subtly. We record
them so that they are not re-proposed.

\paragraph{(i) The zero-prior limit.}
Sweeping prior strength and calling $\lim_{\lambda \to 0}\mathrm{Var}[u \mid y]$ the
irreducible ambiguity. With sparse sensors this limit diverges, so the intended difference
of variances is a difference of infinities.

\paragraph{(ii) Fraction of prior variance removed.}
$\rho(x) = 1 - \mathrm{Var}_{\mathrm{post}}(x)/C(x,x)$. For a stationary prior $C(x,x)$ is
constant, so $\rho$ is an affine function of the posterior variance map and carries no
information the variance map does not already have. Our two-panel version of this
diagnostic, which displayed the two as though they were independent views, had panels
agreeing to four decimal places.

\paragraph{(iii) Elasticity under prior rescaling.}
$f(x) = \partial \log \mathrm{Var}_{\mathrm{post}}(x)/\partial \log \kappa^2$ for a prior
scaled by $\kappa^2$. This one is subtle, because it is well defined and we verified it to
$1.7\times10^{-4}$ against finite differences before noticing the problem. In the eigenbasis
of the prior-preconditioned Hessian it weights direction $i$ by $1/(1+\lambda_i)$, which is
near zero exactly where $\lambda_i$ is large, that is, exactly in the well-informed
directions one is trying to see; at $\lambda \approx 4\times10^{4}$ the weight is
$2\times10^{-5}$. On a planted test with a deliberately inhomogeneous information
distribution it produced $1.1\%$ spatial contrast. It measures where the prior still
dominates, which is the complement of the question.

\section{Certification details}
\label{app:cert}

Backward heat with $N = 128$, $\nu = 5\times10^{-3}$, $t_{\mathrm{obs}} = 1$, $12$ sensors
in a half-domain layout, $\sigma_y = 10^{-2}$, and prior parameters $\alpha = 1$,
$\tau = 8$. The posterior is Gaussian in closed form, giving $\dofref = 5.971$ exactly.
Table~\ref{tab:cert} reports unbiasedness and the convergence rate.

\begin{table}[ht]
\centering
\small
\caption{Certification of the black-box estimator. Backward heat, $N=128$, $n=12$, exact
posterior, $\dofref = 5.971$; $40$ independent probe sets per row. From $P=8$ to $P=128$
the per-run standard deviation falls by a factor $5.0$ against the $4.0$ predicted by
$1/\sqrt{P}$.}
\label{tab:cert}
\begin{tabular}{rrrrr}
\toprule
$P$ (probes) & solves & bias & sd of mean & per-run sd \\
\midrule
8   & 17  & $-0.094$ & 0.142 & 0.897 \\
32  & 65  & $-0.011$ & 0.070 & 0.442 \\
128 & 257 & $-0.009$ & 0.028 & 0.179 \\
\bottomrule
\end{tabular}
\end{table}

From $P = 8$ to $P = 128$, a $16\times$ increase, the per-run standard deviation falls
$0.897 \to 0.179$ and the standard deviation of the mean falls $0.142 \to 0.028$, both
factors of $5.0$ against the $4.0$ predicted by $1/\sqrt{P}$. On the stochastic sampler,
preconditioned unadjusted Langevin returns $6.33 \pm 0.59$ against the exact $5.97$.

\paragraph{Sampling infrastructure.}
Prior-preconditioned sampling gave an effective sample size of about $40$ out of $150$k
steps with under-dispersion of $2$ to $8\times$ across all schemes, exact ones included.
Low-rank posterior preconditioning \citep{flath2011lowrank,spantini2015optimal} at rank
equal to the sensor count, which is exact because point data inform at most $n$ directions,
raised the unadjusted-Langevin stability threshold from $9.7\times10^{-5}$ to $2.0$ and the
effective sample size from $40$ to $13{,}753$, after which the sampler matched the
closed-form posterior to $0.3\%$ in mean and $0.6\%$ in standard deviation.

\section{Full guidance-weight sweep}
\label{app:sweep}

The sweep behind Figure~\ref{fig:localization}(b). Backward heat with $64$ field unknowns
and $8$ sensors, $\dofref = 7.848$, $\epsilon = \sigma_y$, $P = 24$ probes, ensembles of
$512$, averaged over $4$ realisations with $\pm$ the across-realisation standard error.

\begin{table}[ht]
\centering\small
\begin{tabular}{rrrr}
\toprule
$\zeta$ & $\dofsamp$ & stderr & $\rhores$ \\
\midrule
$0.1$ & 1.755 & 0.103 & 0.224 \\
$0.2$ & 4.028 & 0.370 & 0.513 \\
$0.3$ & 6.247 & 0.300 & 0.796 \\
$0.5$ & 7.231 & 0.575 & 0.921 \\
$0.7$ & 8.405 & 0.628 & 1.071 \\
$1$   & 9.565 & 0.363 & 1.219 \\
$1.5$ & 7.532 & 1.859 & 0.960 \\
$2$   & 9.945 & 1.007 & \textbf{1.267} \\
$3$   & 8.266 & 1.015 & 1.053 \\
$5$   & 3.861 & 1.267 & 0.492 \\
$10$  & 1.165 & 5.077 & \textbf{0.148} \\
\bottomrule
\end{tabular}
\end{table}

Two features matter. The curve is not monotone in $\zeta$: it rises to a peak near
$\zeta = 2$ and falls back, so $\dofsamp = \dofref$ has at least two roots and matching
resolution does not identify a weight. And the estimate degrades at the top end, where the
ensemble is over-dispersed by two orders of magnitude (Table~\ref{tab:1d}) and the
across-realisation spread grows accordingly; the $\zeta = 10$ entry has a standard error
larger than its value. We therefore quote the range as a ratio, $0.15$ to $1.27$ times
$\dofref$, rather than as an endpoint-to-endpoint multiplier, which would rest on the
noisiest entry in the table.

\paragraph{Guidance weights are real configuration choices.}
The weights swept here are not a strawman. DiffusionPDE's shipped Darcy configuration uses
three hand-tuned guidance weights spanning three orders of magnitude, differing per PDE
family and absent from that paper's result tables~\citep{huang2024diffusionpde}.

\section{Additional breadth and stability checks}
\label{app:extra}

\paragraph{A non-PDE compressed-sensing inverse problem.}
To check that the measurement is not tied to PDE forward maps, we repeated the controlled
experiment with random compressed linear observations,
$y = Au+\eta$, where $u\in\mathbb{R}^{64}$ has the same exact Gaussian prior,
$A\in\mathbb{R}^{16\times64}$ has independent Gaussian rows normalised to unit norm, and
$\sigma_y=0.05$. This is not a new modelling benchmark; it is a breadth check in which the
prior, diffused score, denoiser and posterior reference all remain exact. The warranted
resolution is $\dofref=15.78$, close to the sensor count because the random measurements
are informative under this prior.

\begin{table}[ht]
\centering\small
\caption{Non-PDE breadth check: random compressed sensing with an exact prior, exact score
and exact posterior reference. Resolution is estimated with $\epsilon=\sigma_y$, $P=24$
probes, ensembles of $512$, and $4$ independent realisations. The exact-score control again
recovers $\dofref$, while distributional spread and resolution fidelity separate: projection
has nearly correct $\rhores$ but about $2.1\times$ posterior variance, and DPS can have
near-reference $\rhores$ with much larger spread.}
\label{tab:compressed}
\begin{tabular}{lrrrrr}
\toprule
guidance & mean err & var ratio & CRPS & cov@90\% & $\rhores$ \\
\midrule
exact score      & 0.102 & 0.96 & 0.271 & 0.812 & 0.991 {\scriptsize$\pm$.001} \\
$\Pi$GDM         & 0.246 & 1.17 & 0.283 & 0.828 & 0.899 {\scriptsize$\pm$.004} \\
projection       & 0.603 & 2.11 & 0.484 & 0.797 & 1.014 {\scriptsize$\pm$.000} \\
DPS $\zeta=0.3$ & 0.697 & 21.4 & 0.521 & 0.922 & 1.023 {\scriptsize$\pm$.064} \\
DPS $\zeta=1$   & 0.409 & 17.0 & 0.402 & 1.000 & 1.079 {\scriptsize$\pm$.091} \\
DPS $\zeta=3$   & 0.126 & 33.4 & 0.393 & 1.000 & 0.839 {\scriptsize$\pm$.066} \\
\bottomrule
\end{tabular}
\end{table}

The compressed-sensing result is deliberately modest and should not be read as a third
application benchmark. It does not address released checkpoints, independent codebases or
image-domain inverse solvers, but it does remove the most immediate concern that the
measurement only behaves on PDE operators. The same pattern recurs: exact guidance is
certified against the reference, and $\rhores$ provides information not contained in CRPS,
coverage or posterior variance.

\paragraph{Higher-replicate stability stress test.}
We also repeated the scalar 1D measurement for the rows that drive
Table~\ref{tab:1d}, using $6$ independent realisations and $P=12$ probes per realisation.
This is intentionally a stress test with half the probe budget of the main table, so the
numbers below are not used as headline estimates. They confirm the control arm and reveal
the finite-budget failure mode: once DPS becomes highly over-dispersed, across-realisation
variation dominates the probe standard error.

\begin{table}[ht]
\centering\small
\caption{Higher-replicate 1D stability stress test, with $\dofref=7.85$,
$\epsilon=\sigma_y$, $P=12$ probes, ensembles of $512$, and $6$ realisations. The exact
control remains stable. Low-weight DPS remains below the warranted resolution, while
larger weights are noisy at this reduced probe budget, consistent with the unstable
high-$\zeta$ maps in Figure~\ref{fig:localization}.}
\label{tab:stability}
\begin{tabular}{lrrrr}
\toprule
guidance & $\dofsamp$ & $\rhores$ & sd$(\rhores)$ & median probe se \\
\midrule
exact score      & $7.77\pm0.03$ & 0.990 & 0.009 & 0.039 \\
DPS $\zeta=0.3$ & $5.50\pm0.39$ & 0.701 & 0.123 & 1.128 \\
DPS $\zeta=1$   & $8.92\pm1.94$ & 1.137 & 0.606 & 3.743 \\
DPS $\zeta=3$   & $6.23\pm4.32$ & 0.793 & 1.347 & 10.282 \\
\bottomrule
\end{tabular}
\end{table}

The important point is not the shifted point estimates at reduced $P$, but the scale of the
uncertainty. Exact guidance remains repeatable. DPS at large guidance weights does not: the
same row that gives $97.7\times$ posterior variance in Table~\ref{tab:1d} has realisation
values from $-9.94$ to $17.21$ in this stress test. This supports the paper's practical
recommendation to report perturbation scale, probe budget and realisation variability rather
than a single scalar alone.

\end{document}